\documentclass[10pt,twocolumn,letterpaper]{article}

\usepackage[pagenumbers]{wacv} 
 \usepackage{multirow}
 \usepackage{algorithm}
\usepackage{algorithmic}
\usepackage[pagenumbers]{wacv}
\definecolor{wacvblue}{rgb}{0.21,0.49,0.74}
\usepackage[pagebackref,breaklinks,colorlinks,allcolors=wacvblue]{hyperref}

\def\wacvPaperID{859} 
\def\confName{WACV}
\def\confYear{2027}

\title{Rethinking Gradient Modulation in Multimodal Regression}

\author{
Haojie Yin$^{1}$ \quad
Chengcheng Feng$^{1}$ \quad
Tianyi Liu$^{2}$ \quad
Tianqi Zhang$^{3}$ \quad
Kaizhu Huang$^{1}$\\
$^{1}$Duke Kunshan University, China\\
$^{2}$Xi'an Jiaotong-Liverpool University, China\\
$^{3}$Soochow University, China
}

\begin{document}
\maketitle
\begin{abstract}
Even balanced multimodal learning methods do not consistently translate additional modalities into better regression performance. To understand this limitation, we revisit the optimization mechanism of balanced multimodal learning, using MMPareto as a representative case. We reveal a previously overlooked issue: MMPareto uses a fixed gradient modulation strength throughout training, while different training stages favor different strengths; an inappropriate modulation strength can instead hinder subsequent optimization. To address this issue, we propose SGM, which adapts the modulation strength according to the current training behavior. We provide theoretical and empirical analyses to characterize and validate the modulation decisions made by SGM. We further build a two-stage multimodal regression framework that integrates AM for unimodal regression representation learning and SGM for adaptive multimodal optimization. Extensive experiments under the same total training budget demonstrate consistent improvements over strong unimodal and multimodal baselines. Our code is provided in the supplementary material.
\end{abstract}
    
\section{Introduction}
\label{sec:intro}

Many real-world applications involve data from multiple modalities~\cite{baltruvsaitis2018multimodal}. For example, clinical assessment often combines different medical imaging modalities to support diagnosis~\cite{perrin2009multimodal}, while autonomous driving relies on multiple sensors to perceive the surrounding environment~\cite{cho2014multi}. Effectively exploiting such multimodal information is therefore an important goal of multimodal learning.

Multimodal regression aims to exploit information from multiple modalities to predict continuous-valued targets. To achieve this goal, existing methods~\cite{zadeh-etal-2017-tensor,liu2018efficient,tsai2019multimodal,ma2021trustworthy} have developed various fusion strategies to capture complementary cross-modal information. However, effectively combining multimodal features does not necessarily ensure that all modalities are sufficiently learned during joint training. Studies of multimodal learning have shown that an easier-to-learn modality may dominate the optimization process, leaving other modalities under-optimized and preventing their complementary information from being fully exploited~\cite{peng2022balanced}. This imbalanced multimodal learning problem has motivated a growing line of balanced multimodal learning methods~\cite{peng2022balanced,fan2023pmr}. MMPareto~\cite{wei2024mmpareto}, as a representative method, integrates multimodal and unimodal gradients through Pareto-based optimization, and then multiplies the resulting modality-encoder gradients by a fixed \textit{modulation strength}.

\begin{figure}[t]
\centering
\includegraphics[width=\linewidth]{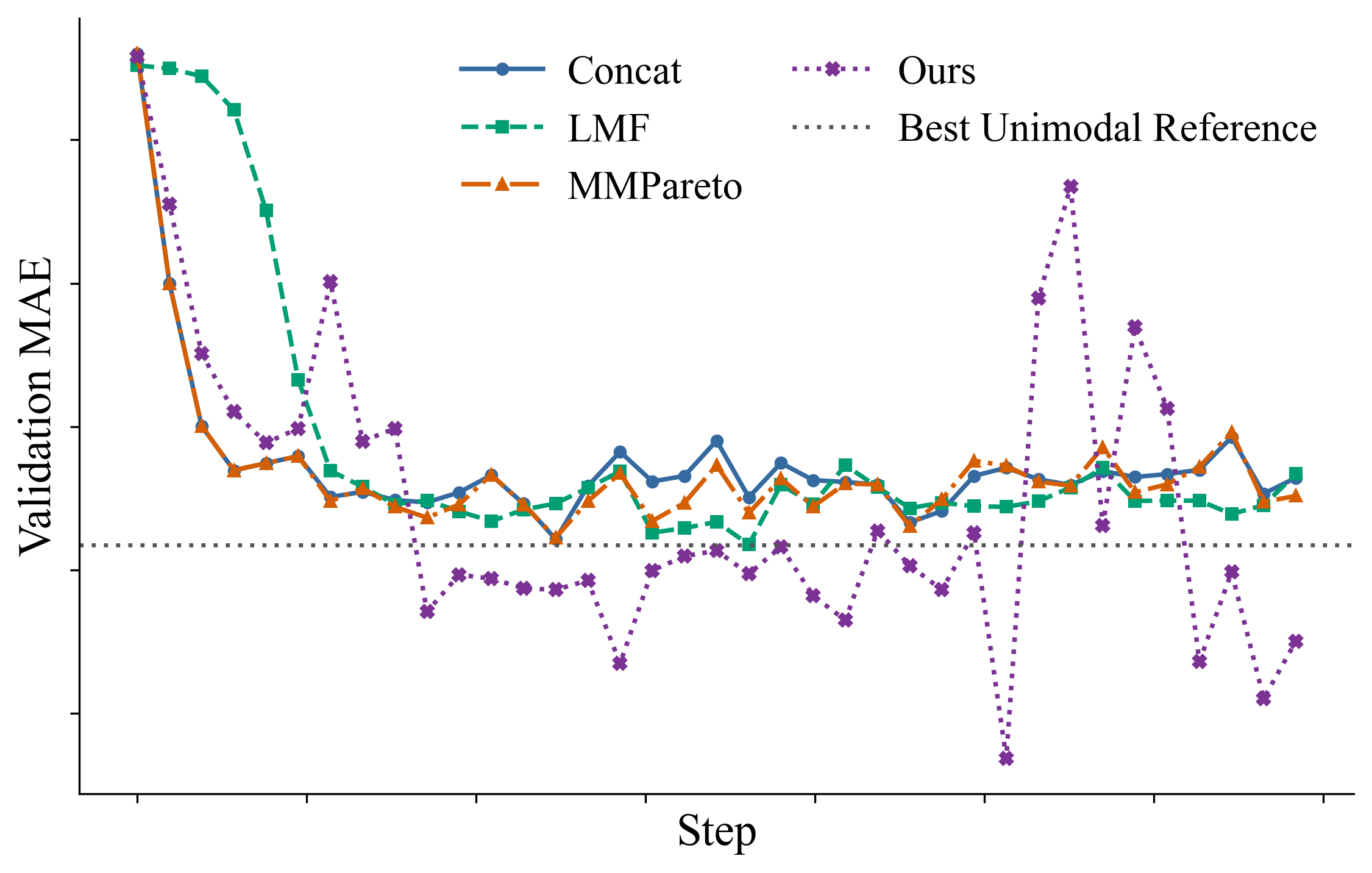}
\caption{Validation performance on STAGE2 over training steps. Multimodal baselines perform comparably to or even worse than the best unimodal model, whereas our method consistently surpasses the unimodal counterpart, indicating more effective exploitation of complementary information across modalities.}
\label{moti}
\end{figure}

However, \textbf{increasingly sophisticated multimodal methods do not necessarily translate additional modalities into better regression performance}. As shown in Fig.~\ref{moti}, on the ophthalmic multimodal regression dataset~\cite{zhou2026stage}, multimodal models perform comparably to, or even worse than, the best unimodal model. This behavior is observed not only with simple concatenation and low-rank fusion (LMF)~\cite{liu2018efficient}, but also with MMPareto~\cite{wei2024mmpareto}. This curious finding serves as the starting point of our work.

To understand why MMPareto does not consistently yield multimodal gains, we further examine its optimization process. We observe that gradient agreement, update-direction stability, and local directional curvature vary markedly across training stages. In other words, the gradients being modulated change substantially throughout training, whereas the modulation strength remains fixed. This motivates us to ask whether different training stages favor different modulation strengths.

Our controlled modulation experiments show that they do. \textbf{A larger modulation strength improves subsequent optimization at some training stages, whereas a smaller modulation strength is preferred at others.} These results indicate that the fixed modulation strength used by MMPareto cannot remain equally suitable throughout training.

We therefore propose \textbf{SGM}, which adaptively adjusts the modulation strength according to the current training behavior. Across the analyzed training stages, SGM changes the modulation strength in the same direction as the choice that yields better subsequent optimization. Importantly, SGM determines the modulation strength online without explicitly searching over multiple candidate strengths.

Our overall framework consists of two stages. In the first stage, we learn unimodal regression representations through contrastive learning. Since continuous labels do not naturally define positive pairs, we treat samples within a label-distance \textit{margin} as positives, where a larger margin provides more positive samples but includes samples with larger label differences, whereas a smaller margin preserves finer label distinctions but provides fewer positives. We therefore introduce \textbf{AM}, which begins with a larger margin to ensure sufficient positive samples and progressively reduces it during training to emphasize finer label distinctions. The resulting unimodal encoders are then used to initialize the second-stage multimodal training with SGM.

Our contributions are three-fold:
\begin{enumerate}

    \item We identify a previously overlooked issue in balanced multimodal learning: the mismatch between fixed gradient modulation and the evolving multimodal optimization. Since the modulation strength directly controls the magnitude of modality-encoder updates, fixing it throughout training can hinder effective multimodal optimization.

    \item We systematically investigate how the appropriate modulation strength varies across training stages and, based on this analysis, propose SGM, which adaptively determines the modulation strength according to the current training behavior. We further provide theoretical and empirical analyses to characterize and validate the modulation choices made by SGM.

    \item We develop a two-stage multimodal regression framework that combines AM for unimodal representation learning with SGM for adaptive multimodal optimization. Extensive experiments demonstrate the effectiveness and robustness of our framework under the same total training budget as competing methods, consistently improving multimodal regression performance over strong unimodal and multimodal baselines.

\end{enumerate}

\section{Related Work}

\subsection{Contrastive Learning for Regression}

Contrastive learning has been increasingly explored for continuous regression tasks~\cite{li2024anchored,trauble2024contrastive}.
Unlike classification, regression labels do not naturally define discrete positive and negative pairs.
Existing methods therefore exploit label proximity to construct contrastive relations between samples.
For example, EchoMEN~\cite{lai2024echomen} defines positive pairs according to label distance and adjusts sample separation based on their relative distances.
However, these methods typically employ a fixed rule for constructing positive pairs throughout training.
Such a fixed neighborhood may be overly coarse once the representation becomes more refined, as nearby samples with different continuous targets may still need to be distinguished.
This motivates an adaptive label-distance margin that gradually shifts from providing sufficient positive pairs to preserving finer label distinctions during training.

\subsection{Multimodal Learning Imbalance}

The simplest way to address learning imbalance is to control the fitting speed of different modalities, giving the harder modality sufficient time to converge.
OGM-GE introduces separate output heads for each modality to monitor their individual learning states and down-weights the dominant unimodal gradient, thereby synchronizing learning speeds~\cite{peng2022balanced}.
MMPareto further improves this by adopting Pareto optimality from multi-task learning, encouraging multimodal and unimodal gradients toward a Pareto-optimal solution~\cite{wei2024mmpareto}.
This promotes better utilization of different modalities and mitigates optimization conflicts.

\begin{figure*}[t]
\centering
\includegraphics[width=0.8\linewidth]{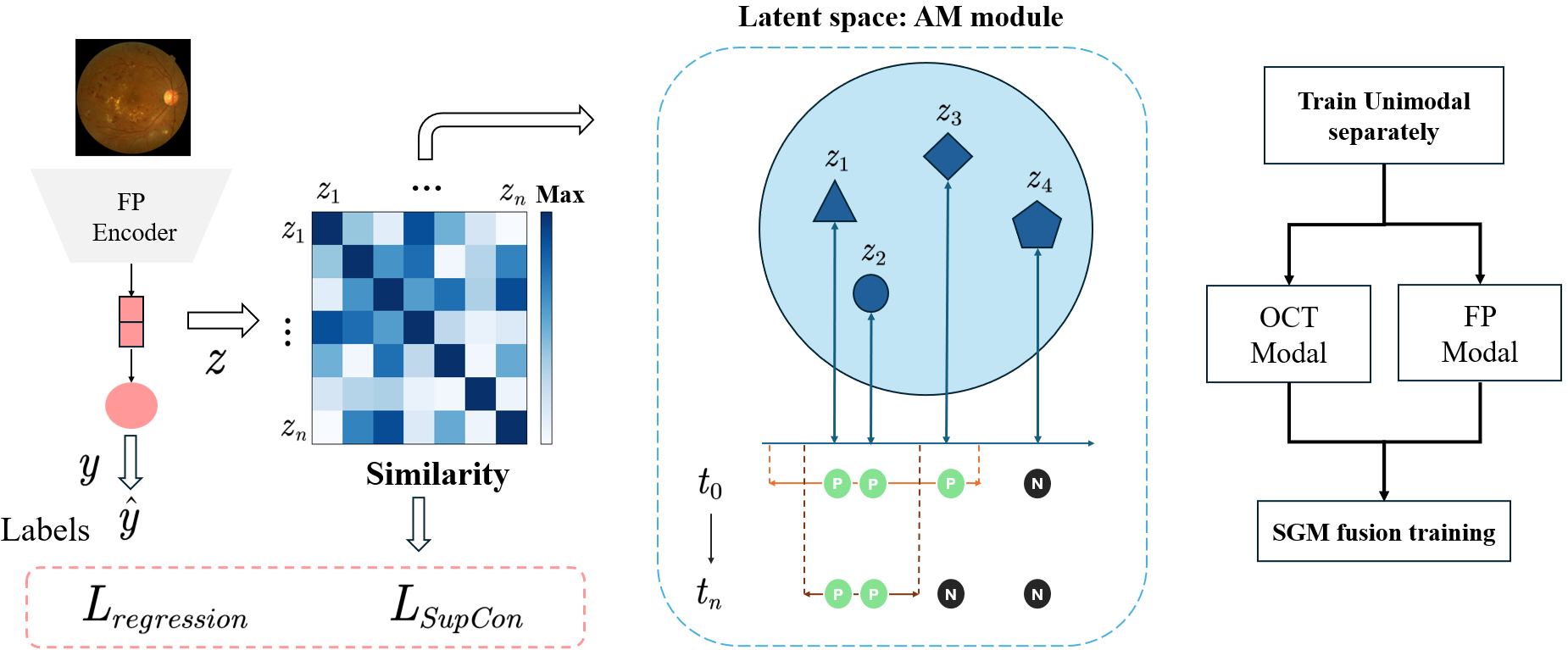}
\caption{Illustration of adaptive margin in supervised contrastive learning. P: positive; N: negative. Early in training, $z_1$ and $z_3$ are positives for $z_2$, while $z_4$ is negative. As training progresses, the margin shrinks, and distant positive $z_3$ becomes negative.}
\label{am}
\end{figure*}

\section{Method}

We consider a multimodal regression dataset
$\mathcal{D}=\{(x_i,y_i)\}_{i=1}^{N}$,
where $x_i=\{x_i^{m}\}_{m=1}^{M}$ contains observations from $M$ modalities and $y_i\in\mathbb{R}$ is the continuous target.
Our goal is to learn a multimodal regressor that predicts $\hat{y}_i\in\mathbb{R}$ from $x_i$.

Our framework consists of two training stages.
In Stage~1, each unimodal encoder is trained independently through regression and contrastive learning.
AM progressively adjusts the label-distance margin used to construct positive pairs.
In Stage~2, the pretrained modality encoders are jointly optimized for multimodal regression.
For each modality encoder, MMPareto~\cite{wei2024mmpareto} integrates the gradients of the multimodal and corresponding unimodal objectives into a common update direction.
While MMPareto scales the integrated gradient using a fixed modulation strength, SGM adaptively determines this strength during training.

\subsection{Stage~1: AM for Unimodal Regression Representation Learning}

In Stage~1, we independently train each unimodal encoder through regression and supervised contrastive learning.
Since continuous labels do not naturally define positive pairs, we regard samples within a label-distance margin as positives.
A larger margin provides more positive samples but includes samples with larger label differences, whereas a smaller margin preserves finer label distinctions but provides fewer positives.
We therefore introduce \textbf{AM}, which starts from a larger margin and progressively reduces it during training:
\begin{equation}
m(t)=
\begin{cases}
m_0, & t<t_n,\\
m_0\exp[-\beta(t-t_n)], & t\geq t_n,
\end{cases}
\end{equation}
where $m_0$, $t_n$, and $\beta$ denote the initial margin, warm-up duration, and decay rate, respectively.

For anchor $i$, the positive set is
\begin{equation}
\mathcal{P}_i(t)=
\{j\neq i:\ |y_i-y_j|<m(t)\}.
\end{equation}
Using these positives, each modality is optimized with
\begin{equation}
\mathcal{L}_{\mathrm{Stage1}}
=
\mathcal{L}_{\mathrm{reg}}
+
\lambda_{\mathrm{con}}\mathcal{L}_{\mathrm{SupCon}},
\end{equation}
where $\mathcal{L}_{\mathrm{SupCon}}$ is the standard supervised contrastive loss~\cite{khosla2020supervised} constructed with $\mathcal{P}_i(t)$.
The resulting unimodal encoders initialize the multimodal model in Stage~2.

\begin{figure}[t]
\centering
\includegraphics[width=\linewidth]{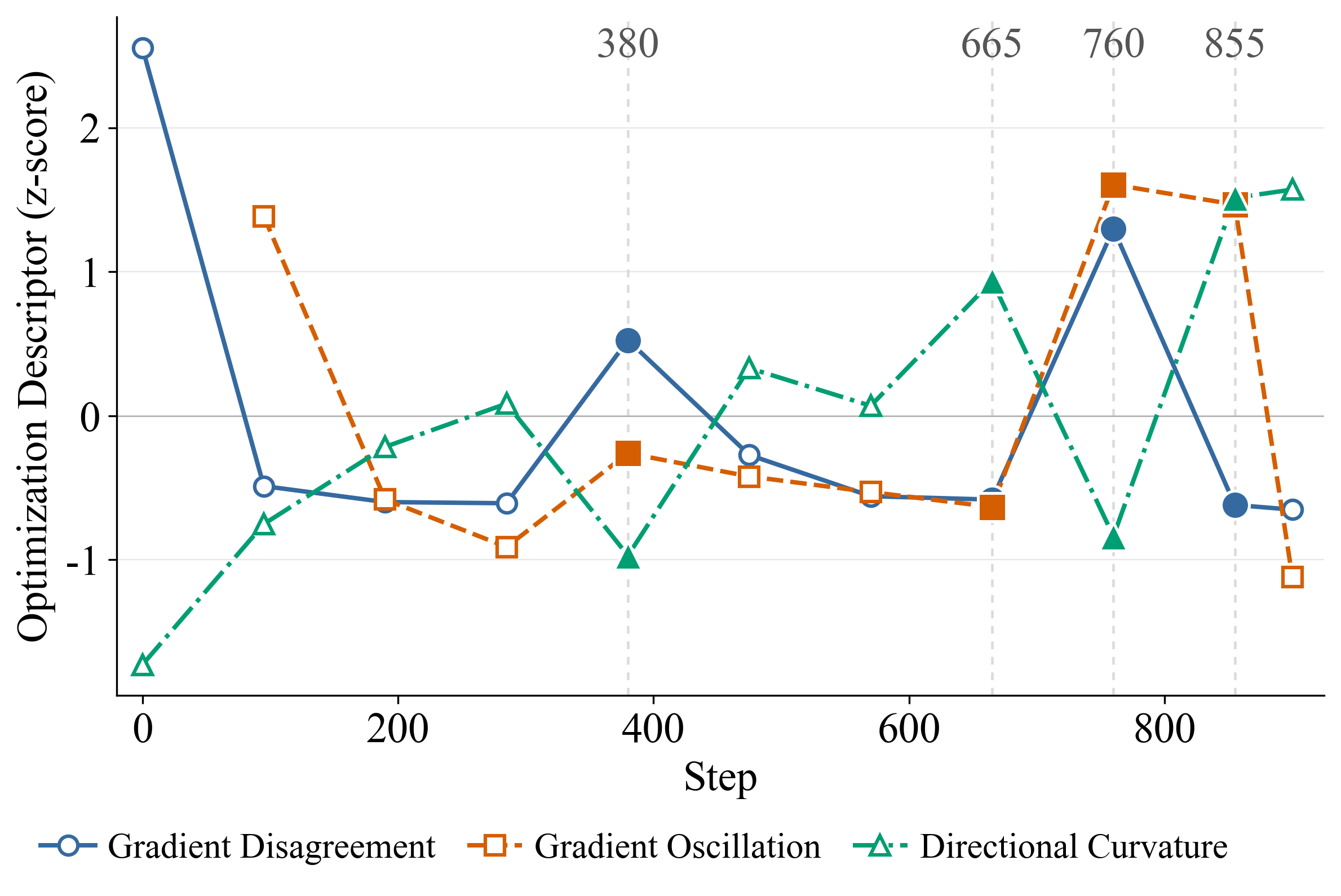}
\caption{Optimization conditions evolve differently during multimodal training. To compare gradient disagreement, gradient oscillation, and directional curvature on a common scale, each descriptor is independently standardized using its trajectory-wise z-score, with zero denoting its own mean level.}
\label{opti}
\end{figure}

\begin{figure}[t]
\centering
\includegraphics[width=\linewidth]{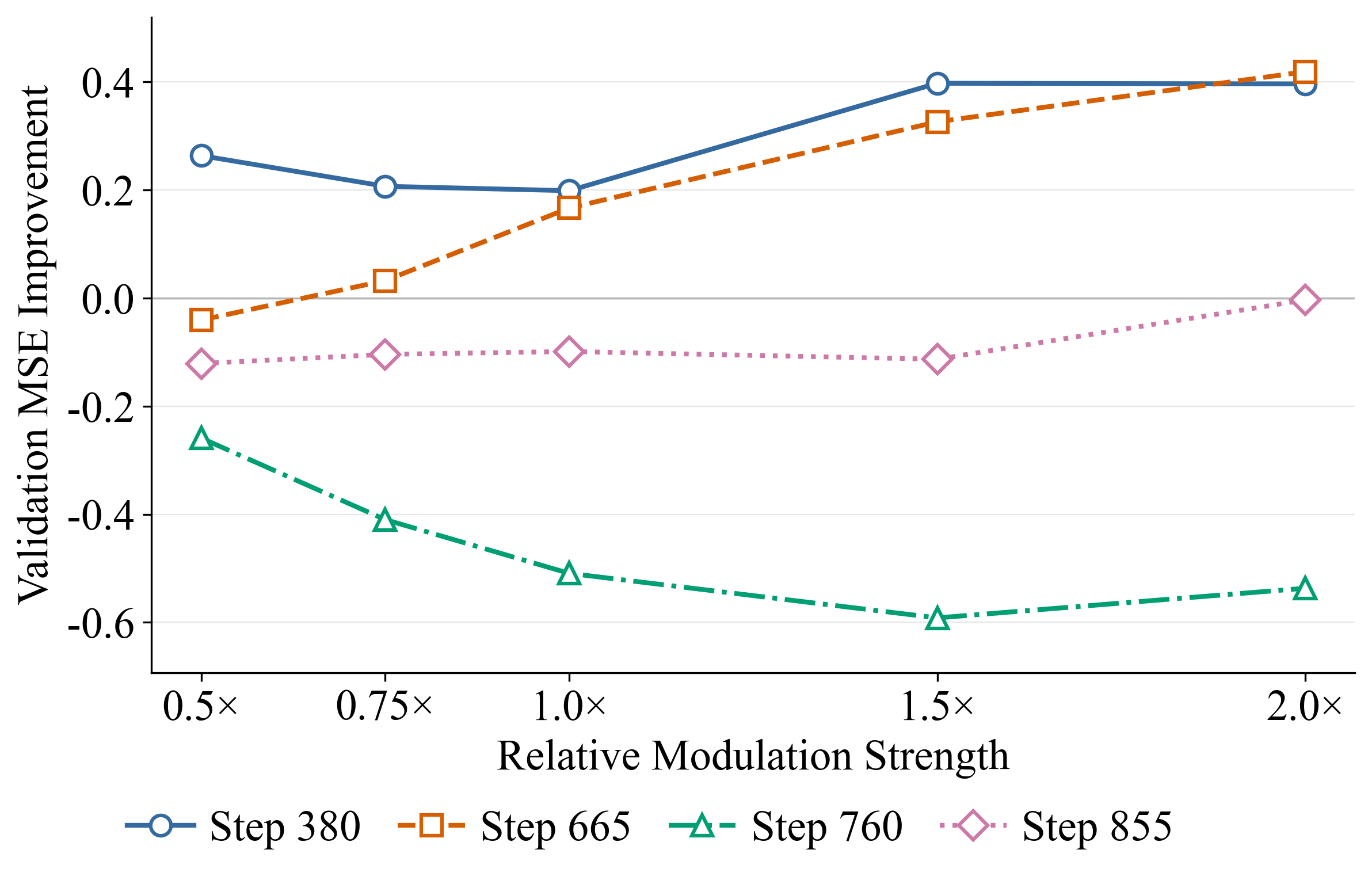}
\caption{Different Training Checkpoints Prefer Different Modulation Strengths. The x-axis denotes the modulation strength relative to the fixed baseline, where \(1.0\times\) corresponds to the original modulation strength. The y-axis denotes the reduction in validation MSE from the source checkpoint after a fixed optimization horizon, where a larger value indicates greater improvement and a negative value indicates degraded validation performance.}
\label{modu}
\end{figure}

\subsection{Diagnosing Fixed Gradient Modulation}

We next examine whether the fixed modulation strength in MMPareto remains suitable throughout training.
For each modality encoder, MMPareto integrates the gradients from the multimodal and corresponding unimodal objectives into an update $\tilde{\mathbf g}_t$, and scales its magnitude using a fixed modulation strength $\gamma$:
\begin{equation}
    \mathbf g_t^{\mathrm{mod}}
    =
    \gamma \tilde{\mathbf g}_t .
\end{equation}
Although $\tilde{\mathbf g}_t$ is determined by the gradients at the current training step, the same $\gamma$ is used throughout training.

We first inspect how the optimization behavior changes over time. As shown in Fig.~\ref{opti}, we monitor three complementary quantities: gradient disagreement between multimodal and unimodal objectives, gradient oscillation measured by the change in update direction between consecutive steps, and local directional curvature. These quantities exhibit markedly different patterns across training stages. For example, Step 665 shows low gradient disagreement and gradient oscillation but high directional curvature, whereas Step 760 exhibits high gradient disagreement and gradient oscillation with much lower directional curvature. Such contrasting optimization behaviors motivate us to investigate whether different training stages favor different modulation strengths.

To answer this question, we perform controlled modulation experiments from several training checkpoints. For each checkpoint, we branch from the same model state using relative modulation strengths ${0.5,0.75,1.0,1.5,2.0}\gamma$, while keeping all other training conditions unchanged. We evaluate each branch by the reduction in validation MSE from the source checkpoint after a fixed optimization horizon, where a larger value indicates greater improvement. As shown in Fig.~\ref{modu}, different checkpoints favor different modulation strengths. For example, at Step~665, increasing the modulation strength consistently improves subsequent validation MSE, whereas at Step~760, stronger modulation leads to worse subsequent optimization and a smaller strength is preferred.

These results show that a single fixed modulation strength cannot remain equally suitable throughout training.
Instead, the modulation strength should adapt to the current training condition, which motivates the SGM strategy introduced next.
\begin{figure*}[t]
\centering
\includegraphics[width=1\linewidth]{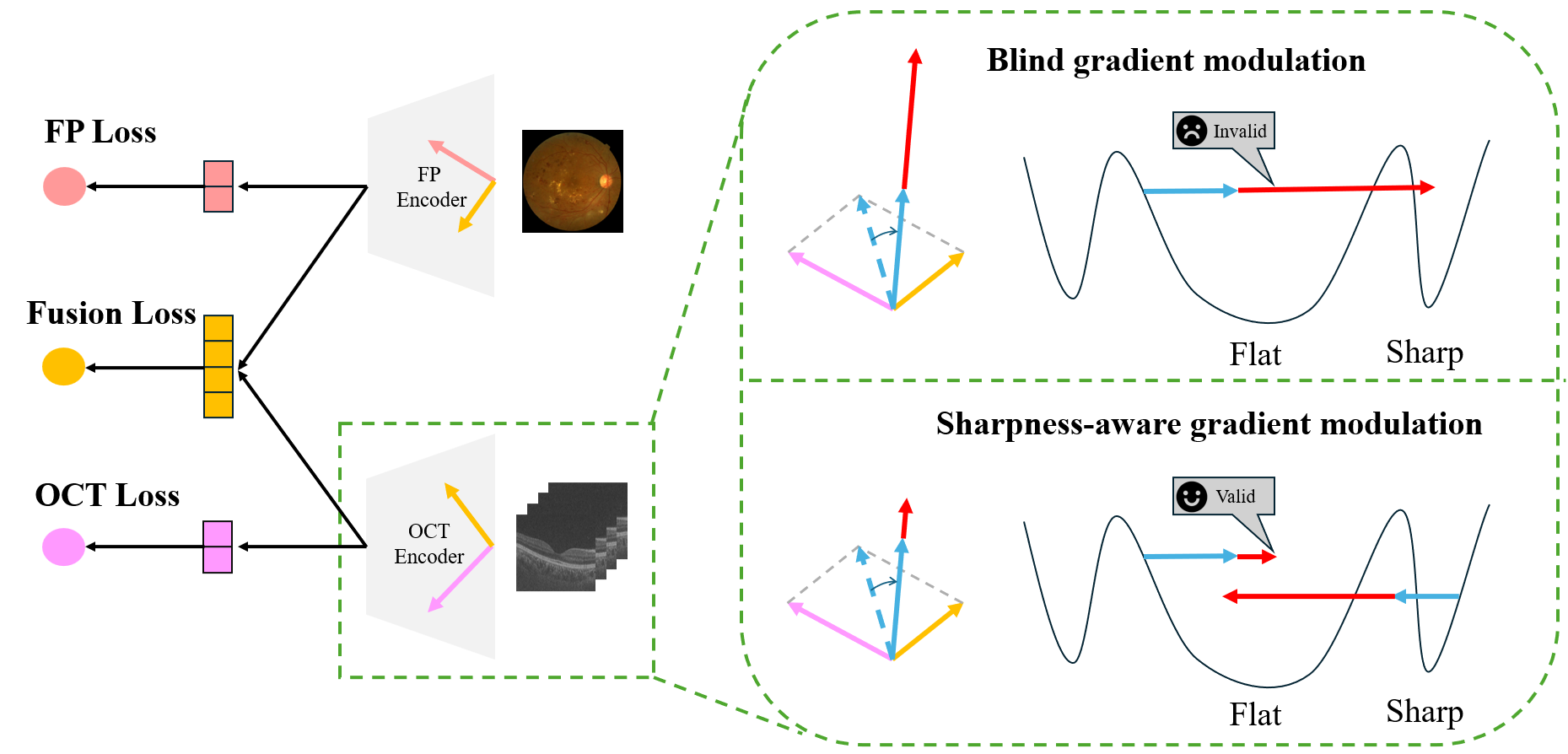}
\caption{Illustration of SGM on the STAGE2 ophthalmic multimodal regression task. Taking the OCT encoder as an example, MMPareto integrates gradients from the multimodal and OCT objectives into a common update direction and scales the resulting gradient using a fixed modulation strength. SGM instead adapts the modulation strength according to the current training behavior.}
\label{sgm}
\end{figure*}

\subsection{Stage~2: SGM for Adaptive Gradient Modulation}

We propose SGM to adapt the modulation strength according to the current training behavior.
Rather than explicitly evaluating multiple candidate strengths, SGM constructs a lightweight signal from the current loss and its recent history.

Let $\mathcal{L}_{\mathrm{mm}}$ denote the multimodal regression loss and
$\mathcal{L}_{m}$ the unimodal loss associated with modality $m$.
We define the objective used for modulation estimation as
\begin{equation}
    \mathcal{L}_{\mathrm{s}}
    =
    \mathcal{L}_{\mathrm{mm}}
    +
    \sum_{m=1}^{M}\mathcal{L}_{m}.
\end{equation}
At training step $t$, we compute
$\mathbf{g}_{t}=\nabla_{\theta}\mathcal{L}_{\mathrm{s}}(\theta_t)$
and construct a small gradient-aligned perturbation for each parameter tensor $p$:
\begin{equation}
    \delta_{t,p}
    =
    \epsilon
    \frac{\mathbf{g}_{t,p}}
    {\|\mathbf{g}_{t,p}\|_2+\varepsilon},
\end{equation}
where $\epsilon$ controls the perturbation scale and $\varepsilon$ is a small constant for numerical stability.
We then measure the local loss response along this direction by
\begin{equation}
    s_t
    =
    \mathcal{L}_{\mathrm{s}}(\theta_t+\delta_t)
    -
    \mathcal{L}_{\mathrm{s}}(\theta_t-\delta_t).
\end{equation}
A larger $s_t$ indicates a stronger loss variation around the current parameters along the gradient-related direction.

Since the absolute scale of $s_t$ can change during training, we compare the current response with its recent history.
Let $W_t$ denote a sliding window containing the recent response values.
The relative modulation signal is defined as
\begin{equation}
    q_t
    =
    \frac{s_t}
    {\operatorname{median}(W_t)+\varepsilon}.
\end{equation}
Thus, $q_t$ measures the current loss response relative to its recent reference level rather than relying on its absolute magnitude.

SGM uses this relative signal to adapt the modulation strength:
\begin{equation}
    \gamma_t
    =
    \operatorname{clip}
    \left(
    \gamma_{\mathrm{base}} q_t,
    \gamma_{\min},
    \gamma_{\max}
    \right),
\end{equation}
where $\gamma_{\mathrm{base}}$ is the original modulation strength and
$\gamma_{\min}$ and $\gamma_{\max}$ bound its allowable range.
Before sufficient response history is collected, we use
$\gamma_t=\gamma_{\mathrm{base}}$.

For each modality encoder, let
$\tilde{\mathbf{g}}_{t}^{m}$
denote the gradient obtained after MMPareto integrates the multimodal and corresponding unimodal gradients.
SGM replaces the fixed modulation strength with $\gamma_t$:
\begin{equation}
    \mathbf{g}_{t}^{m,\mathrm{SGM}}
    =
    \gamma_t
    \tilde{\mathbf{g}}_{t}^{m}.
\end{equation}

The resulting gradients are then passed to the optimizer for parameter updates, while parameters outside the modulated modality-encoder components follow the original optimization procedure.

In this way, SGM adjusts the gradient modulation strength directly during training without explicitly branching over multiple candidate values.
We theoretically analyze how changes in the modulation strength affect the subsequent optimizer update in Appendix, and empirically examine whether the modulation choices made by SGM agree with better subsequent optimization in Sec.~\ref{sec:sgm_validation}.

\begin{table}[]
\centering
\caption{Comparison of performance on the STAGE2 dataset. Results are reported as the mean over four random seeds.}
\label{tab:stage2_results}
\begin{tabular}{lcccc}
\toprule
\textbf{Methods} 
& \textbf{MAE $\downarrow$}
& \textbf{MSE $\downarrow$}
& \textbf{SMAPE $\downarrow$}
& $\mathbf{R^2 \uparrow}$ \\
\midrule
OCT-only & 1.589 & 4.735 & 0.640 & 0.257 \\
FP-only  & 1.407 & 3.501 & 0.653 & 0.435 \\
\midrule
MMPareto & 1.400 & 4.112 & 0.620 & 0.477 \\
DGL      & 1.376 & 3.127 & 0.630 & 0.487 \\
RGM      & 1.524 & 4.258 & 0.651 & 0.276 \\
M-SAM    & 1.599 & 4.785 & 0.659 & 0.316 \\
MASAM    & 1.364 & 3.436 & \textbf{0.614} & 0.445 \\
G2D      & 1.507 & 3.662 & 0.645 & 0.307 \\
\midrule
\textbf{Ours} 
& \textbf{1.306} 
& \textbf{3.011} 
& 0.633 
& \textbf{0.506} \\
\bottomrule
\end{tabular}
\end{table}

\begin{table*}[t]
\centering
\caption{Comparison of performance on CMU-MOSI and CH-SIMS.
Results are reported as the mean over four random seeds.
$\downarrow$ indicates lower is better, while $\uparrow$ indicates higher is better.
The best results are highlighted in \textbf{bold}.}
\label{tab:main_regression}
\setlength{\tabcolsep}{4.5pt}
\renewcommand{\arraystretch}{1.15}

{
\begin{tabular}{l||cccc||cccc}
\toprule
\multirow{2}{*}{\textbf{Methods}}
&
\multicolumn{4}{c||}{\textbf{CMU-MOSI}}
&
\multicolumn{4}{c}{\textbf{CH-SIMS}}
\\

&
\textbf{MAE} $\downarrow$
&
\textbf{MSE} $\downarrow$
&
\textbf{SMAPE} $\downarrow$
&
$\mathbf{R^2}\uparrow$
&
\textbf{MAE} $\downarrow$
&
\textbf{MSE} $\downarrow$
&
\textbf{SMAPE} $\downarrow$
&
$\mathbf{R^2}\uparrow$
\\

\midrule

Text-only
& $0.956$
& $1.511$
& $0.994$
& $0.400$
& $0.443$
& $0.349$
& $\mathbf{1.014}$
& $0.268$
\\

Audio-only
& $1.415$
& $2.683$
& $1.639$
& $-0.063$
& $0.564$
& $0.458$
& $1.484$
& $0.041$
\\

Vision-only
& $1.430$
& $2.798$
& $1.572$
& $-0.109$
& $0.506$
& $0.362$
& $1.203$
& $0.241$
\\

\midrule

MMPareto
& $0.967$
& $1.586$
& $0.943$
& $0.372$
& $0.434$
& $0.335$
& $1.033$
& $0.299$
\\

DGL
& $1.104$
& $1.886$
& $1.105$
& $0.252$
& $0.576$
& $0.590$
& $1.153$
& $-0.237$
\\

RGM
& $0.977$
& $1.601$
& $0.953$
& $0.366$
& $0.444$
& $0.341$
& $1.049$
& $0.285$
\\

M-SAM
& $0.958$
& $1.566$
& $0.927$
& $0.379$
& $0.436$
& $0.320$
& $1.016$
& $0.330$
\\

MASAM
& $0.962$
& $1.540$
& $0.963$
& $0.390$
& $0.428$
& $0.312$
& $1.034$
& $0.347$
\\

G2D
& $1.066$
& $1.672$
& $1.116$
& $0.337$
& $0.459$
& $0.322$
& $1.103$
& $0.325$
\\

\midrule

\textbf{Ours}
& $\mathbf{0.943}$
& $\mathbf{1.506}$
& $\mathbf{0.922}$
& $\mathbf{0.403}$
& $\mathbf{0.415}$
& $\mathbf{0.291}$
& $\mathbf{1.014}$
& $\mathbf{0.391}$
\\

\bottomrule
\end{tabular}
}
\end{table*}


\section{Experiment}

\subsection{Datasets}

We evaluate our method on four multimodal regression benchmarks: \textbf{STAGE2}, \textbf{CMU-MOSI}, \textbf{CH-SIMS}, and \textbf{MIS-ME}. \textbf{STAGE2}~\cite{zhou2026stage} is a multimodal medical regression dataset for visual field mean deviation prediction, containing OCT slices and fundus photographs. \textbf{CMU-MOSI} contains 2,199 monologue video clips with text, visual, and acoustic modalities, annotated with sentiment scores ranging from $-3$ to $3$. \textbf{CH-SIMS} is a Chinese multimodal sentiment dataset containing 2,281 video clips collected from movies. \textbf{MIS-ME} is a multimodal regression dataset containing raw soil patch images and meteorological data for soil moisture estimation. Detailed experimental settings and results for \textbf{MIS-ME} are provided in the Appendix.

\subsection{Settings}

We took four classic regression metrics: $R^2$, MSE, MAE and SMAPE. All experiments are conducted on a NVIDIA L20 GPU. The batch size is set to 8. We adopt the Adam optimizer. For STAGE2, we exploit ResNet34 as the backbone network, following the~\cite{wei2024mmpareto}. For CMU-MOSI and CH-SIMS, The network consists of a single layer unidirectional LSTM for text encoding, three-layer ReLU MLPs with temporal mean pooling for audio and visual encoding. Additional details about the datasets, baseline description, code and hyperparameter sensitive analysis can be found in the supplementary material.

\subsection{Comparison with SOTA Methods}

\subsubsection{Comparison Baselines.} To comprehensively evaluate the effectiveness of our proposed method, we compare it against a diverse set of multimodal learning baselines, including MMPareto~\cite{wei2024mmpareto}, DGL~\cite{wei2025boosting}, RGM~\cite{gao2026reconcile}, M-SAM~\cite{rajoli2026modality}, MASAM~\cite{chen2026masam}, G2D~\cite{rakib2025g}. 

\begin{figure}[t]
\centering
\includegraphics[width=\linewidth]{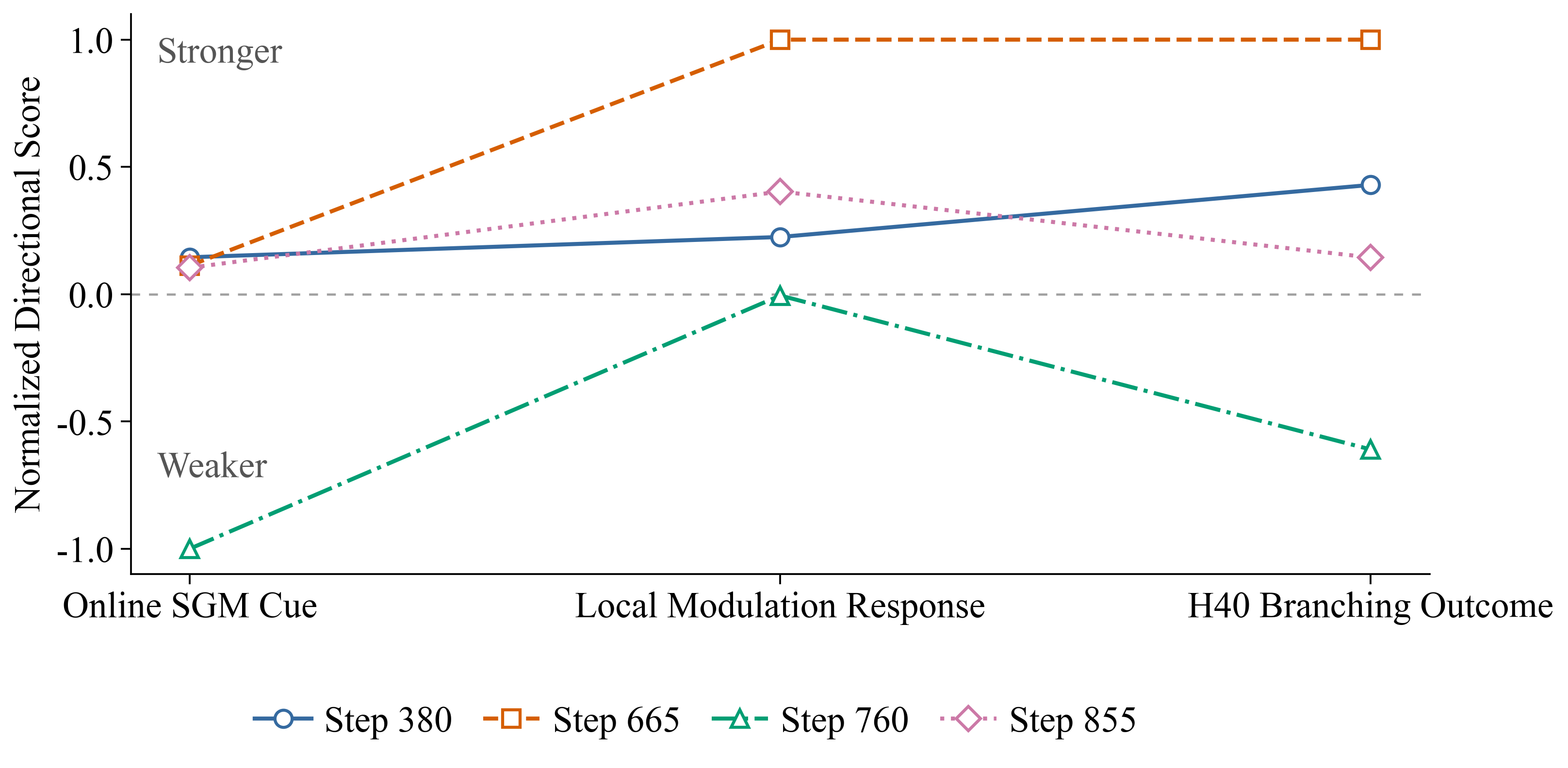}
\caption{Validation of SGM modulation decisions across training stages.
The modulation strength selected by SGM is directionally consistent with that favored by subsequent optimization across all analyzed checkpoints.
Scores are independently normalized for visualization.}
\label{val}
\end{figure}

\subsubsection{Results}

Tables~\ref{tab:stage2_results} and~\ref{tab:main_regression} report the results on the three multimodal regression benchmarks, where all results are averaged over four random seeds. 
On STAGE2, our method achieves the best MAE, MSE, and $R^2$, substantially outperforming both the strongest unimodal model and existing multimodal approaches. 
In particular, it reduces the MSE from 3.501 of the best unimodal model to 3.011, demonstrating that the complementary information from multiple modalities is effectively exploited.

Similar improvements are observed on CMU-MOSI and CH-SIMS. 
Our method achieves the best results across all four metrics on CMU-MOSI, and obtains the best MAE, MSE, and $R^2$ on CH-SIMS while achieving competitive SMAPE. 
These consistent improvements across medical and sentiment regression tasks demonstrate the effectiveness and generalizability of our two-stage framework for multimodal regression. 

Further analyses of the validation experiments for AM and SGM are provided in the Appendix.

\begin{figure}[t]
\centering
\includegraphics[width=\linewidth]{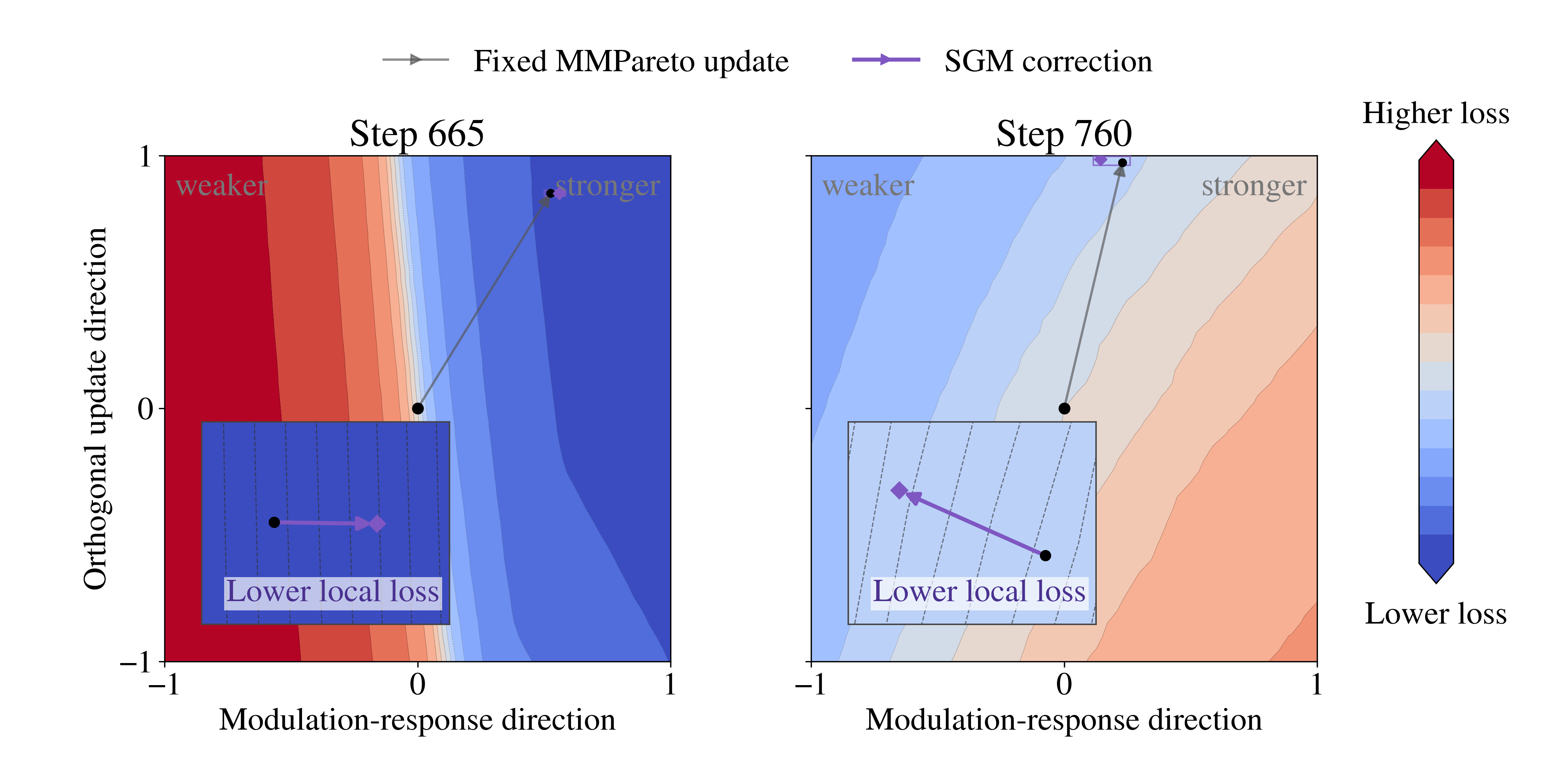}
\caption{Validation of SGM modulation decisions across training stages.
The modulation strength selected by SGM is directionally consistent with that favored by subsequent optimization across all analyzed checkpoints.
Scores are independently normalized for visualization.}
\label{las}
\end{figure}

\subsection{Further Analysis}

\subsubsection{Does SGM Select Appropriate Modulation Strengths?}
\label{sec:sgm_validation}

We further investigate whether the online signal used by SGM provides a meaningful direction for adjusting the modulation strength.
To this end, we select four representative training checkpoints (Steps 380, 665, 760, and 855) and compare the modulation preference from three complementary perspectives, as shown in Fig.~\ref{val}.

First, we consider the \textit{Online SGM Cue}.
Recall that SGM computes the relative loss-response signal
$q_t=s_t/\operatorname{median}(W_t)$
and adjusts the modulation strength accordingly.
We use
\begin{equation}
    Q_t=q_t-1
\end{equation}
to represent its directional preference, where
$Q_t>0$ indicates that SGM favors a larger modulation strength and
$Q_t<0$ indicates that it favors a smaller one.

Second, we independently evaluate the \textit{Local Modulation Response}.
Given the actual optimizer update $\theta_{t+1}(\gamma)$, we measure how a small change in the modulation strength affects the next-step loss:
\begin{equation}
    R_t =
    \frac{\partial
    \mathcal{L}(\theta_{t+1}(\gamma))}
    {\partial \gamma}.
\end{equation}
For consistent interpretation, we plot
$M_t=-R_t$.
Thus, $M_t>0$ means that increasing $\gamma$ locally decreases the next-step loss, whereas $M_t<0$ indicates that a smaller modulation strength is locally preferred.
The detailed derivation and numerical verification of this response are provided in Appendix.

Third, we examine whether these local preferences agree with subsequent optimization through controlled branching experiments.
Starting from each checkpoint, we continue training with relative modulation strengths
$\{0.5,0.75,1.0,1.5,2.0\}\gamma$
while keeping the model state, subsequent mini-batch sequence, and other optimization settings unchanged.
We evaluate the resulting validation MSE after 40 optimization steps and derive the corresponding \textit{H40 Branching Outcome}, where a positive value indicates a preference for larger modulation strength and a negative value indicates a preference for smaller modulation strength.

For visualization, the three directional scores are independently normalized by their maximum absolute values.
Therefore, Fig.~\ref{val} is intended to compare their directions rather than their absolute magnitudes.
Remarkably, the three independently obtained signals exhibit the same directional preference at all four analyzed checkpoints.
At Steps 380, 665, and 855, SGM favors stronger modulation, which agrees with both the local modulation response and the subsequent branching outcome.
In contrast, all three signals reverse at Step 760 and favor weaker modulation.

These results establish a consistent link between the online decision made by SGM, the local effect of changing the modulation strength, and the modulation strength preferred by subsequent optimization.

\subsubsection{How Does SGM Adapt the Gradient Update?}
\label{landscape}

To further understand how SGM modifies the optimization trajectory, we visualize the local loss landscape around two representative training states, Steps 665 and 760.
For each checkpoint, we construct a two-dimensional plane whose horizontal axis corresponds to the modulation-response direction, i.e., the direction associated with changing the modulation strength, while the vertical axis is an orthogonal update direction.
We evaluate the loss over this local plane and compare the update produced by the fixed modulation strength of MMPareto with the correction introduced by SGM.

As shown in Fig.~\ref{las}, the two checkpoints exhibit opposite modulation preferences.
At Step~665, lower local loss lies toward the stronger-modulation side.
Accordingly, SGM increases the modulation strength and shifts the update further in this direction.
In contrast, at Step~760, the local loss decreases toward the weaker-modulation side.
SGM therefore reduces the modulation strength and corrects the update in the opposite direction.
The enlarged regions in Fig.~\ref{las} further illustrate these local corrections.

This comparison highlights the key behavior of SGM:
it does not apply a uniformly stronger or weaker modulation throughout training.
Instead, it modifies the fixed MMPareto update in different directions according to the current training state.
The opposite corrections at Steps~665 and 760 are consistent with the modulation preferences observed in the controlled branching experiments, providing an intuitive visualization of why adaptive modulation is needed.

\begin{figure*}[!t]
    \centering

    \begin{subfigure}[b]{0.33\linewidth}
        \centering
        \includegraphics[width=\linewidth]{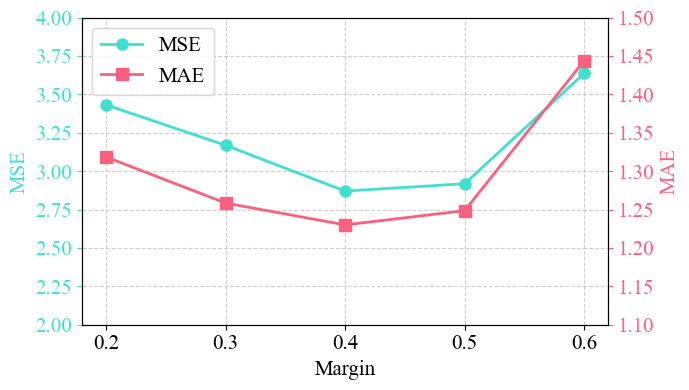}
        \caption{AM: Margin}
        \label{am_m}
    \end{subfigure}
    \hfill
    \begin{subfigure}[b]{0.33\linewidth}
        \centering
        \includegraphics[width=\linewidth]{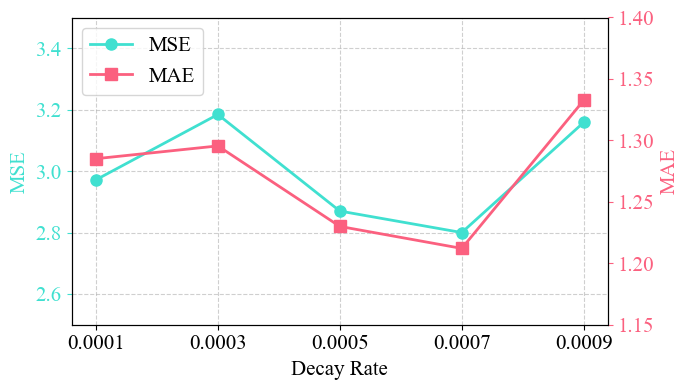}
        \caption{AM: Decay Rate}
        \label{am_d}
    \end{subfigure}
    \hfill
    \begin{subfigure}[b]{0.33\linewidth}
        \centering
        \includegraphics[width=\linewidth]{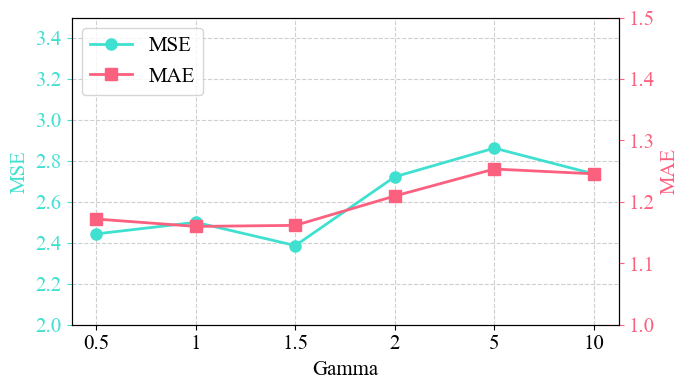}
        \caption{SGM: Gamma}
        \label{sen}
    \end{subfigure}

    \caption{Hyperparameter sensitivity analysis of AM and SGM.}
    \label{fig:sensitivity}
\end{figure*}

\subsection{Ablation Study}

We conduct an ablation study on the STAGE2 dataset to evaluate the contributions of AM and SGM, as reported in Table~\ref{ablation}.
We consider the following variants:
(a) \textbf{w/o AM}, which replaces the adaptive margin in Stage~1 with a fixed margin while retaining SGM in Stage~2;
(b) \textbf{w/o SGM (Unimodal Only)}, which removes Stage~2 multimodal optimization and directly evaluates the unimodal models learned in Stage~1;
(c) \textbf{w/o SGM (Fusion)}, which retains Stage~1 representation learning but replaces SGM with standard joint multimodal optimization; and
(d) \textbf{AM w/ MP}, which retains AM in Stage~1 but uses the original MMPareto with a fixed modulation strength in Stage~2.

Removing AM degrades regression performance, showing the benefit of adaptively constructing positive pairs during unimodal representation learning.
More importantly, replacing SGM with either standard joint optimization or the fixed modulation strategy of MMPareto consistently reduces performance, demonstrating the importance of adaptive gradient modulation in Stage~2.
The complete framework achieves the best overall performance, indicating that AM and SGM provide complementary benefits in the two training stages.

\begin{table}[]
\centering
\small 
\caption{Ablation study on the two-stage framework. “MP” denotes the MMPareto method used in the second stage.}
\label{ablation}
\setlength{\tabcolsep}{1mm}
\begin{tabular}{c|c|cccc}
\toprule
\textbf{Method} & \textbf{Modal} & \textbf{MAE} ↓ & \textbf{MSE} ↓ & \textbf{SMAPE} ↓ & $\boldsymbol{R}^2$ ↑\\
\midrule
Baseline              & Fusion  & 1.52 & 4.64 & 0.629 & 0.371\\
\midrule
w/o AM                 & Fusion  & 1.26 & 2.95 & 0.640 & 0.611\\
\midrule
\multirow{3}{*}{w/o SGM}
                       & OCT     & 1.54 & 4.79 & 0.630 & 0.340 \\
                       & FP      & 1.23 & 2.87 & 0.640 & 0.601 \\
                       & Fusion  & 1.29 & 3.21 & 0.651 & 0.525\\
\midrule
AM w/ MP               & Fusion  & 1.53 & 4.19 & 0.634 & 0.317\\
\midrule
\textbf{Ours}          & Fusion  & \textbf{1.16} & \textbf{2.50} & \textbf{0.581} & \textbf{0.626}\\
\bottomrule
\end{tabular}
\end{table}

\subsection{Hyperparameter Sensitivity.}

We analyze the sensitivity of key hyperparameters in AM and SGM. 
For AM, we vary the initial margin while fixing the decay rate to $0.0005$ (Fig.~\ref{am_m}), and vary the decay rate while fixing the initial margin to $0.4$ (Fig.~\ref{am_d}). 
For SGM, we evaluate the scaling factor $\gamma \in \{0.5, 1, 1.5, 2, 5, 10\}$ (Fig.~\ref{sen}). 
The results show that the performance remains stable across a wide range of hyperparameters, indicating that the proposed method is not sensitive to parameter selection.




\subsection{Cost-Time and Efficiency Analysis}

\begin{table}[h]
\centering
\caption{Cost-time and performance comparison on the STAGE2 dataset. 
All methods are trained with the same total number of epochs.}
\label{cost_time}
\setlength{\tabcolsep}{4mm}
\begin{tabular}{l|cc}
\toprule
\textbf{Method} 
& \textbf{Time / Epoch (s) $\downarrow$} 
& \textbf{MSE $\downarrow$} \\
\midrule
Baseline  & \textbf{385} & 4.640 \\
MMPareto  & 394 & 4.112 \\
DGL       & 403 & 3.127 \\
RGM       & 421 & 4.258 \\
M-SAM     & 414 & 4.785 \\
MASAM     & 420 & 3.436 \\
\midrule
\textbf{Ours} & 404 & \textbf{3.011} \\
\bottomrule
\end{tabular}
\end{table}

We further analyze the computational efficiency of our framework in
Table~\ref{cost_time}. 
Although our method adopts a two-stage training strategy, the total number
of training epochs is kept identical to that of all competing methods.
Therefore, the two-stage design does not introduce an additional training
budget. Moreover, its average training time per epoch (404\,s) remains
comparable to existing multimodal methods, while achieving the lowest MSE
of 3.011.

These results also suggest that effective multimodal learning does not
necessarily require joint optimization of all modalities from the beginning.
Instead, first establishing modality-specific representations and then
performing effective joint optimization can lead to better multimodal
regression performance under the same training budget.
\section{Conclusion}

In this work, we revisit gradient modulation in multimodal regression and identify a previously overlooked limitation of existing balanced multimodal learning: although the optimization state evolves throughout training, the modulation strength is often kept fixed. Through controlled analyses, we show that different training stages can favor substantially different modulation strengths, motivating the need for adaptive modulation. Based on this observation, we introduce SGM, which adjusts the modulation strength online according to the current loss response without explicitly searching over candidate values. We further combine SGM with AM, which progressively refines label-aware contrastive learning for unimodal regression representation learning, resulting in a two-stage multimodal regression framework. Extensive experiments across diverse multimodal regression benchmarks demonstrate that the proposed framework consistently improves over strong unimodal and multimodal baselines under the same total training budget. Further analyses show that the modulation decisions made by SGM are directionally consistent with both local optimization responses and subsequent optimization outcomes. These results highlight the importance of adapting gradient modulation to the evolving optimization state for effectively exploiting complementary information in multimodal regression.

{
    \small
    \bibliographystyle{ieeenat_fullname}
    \bibliography{main}
}


\end{document}